\documentclass[conference]{IEEEtran}
\IEEEoverridecommandlockouts
\usepackage{cite}
\usepackage{authblk}
\usepackage{pgfplots}
\usepackage{tikz}
\usepackage{amsmath,amssymb,amsfonts}
\usepackage{algorithmic}
\usepackage{graphicx}
\usepackage{textcomp}
\usepackage{xcolor}
\usepackage{multirow}
\usepackage{pdfpages}
\usepackage{arydshln}
\usepackage{url} 
\usepgfplotslibrary{polar}
\pgfplotsset{compat=1.16}
\def\BibTeX{{\rm B\kern-.05em{\sc i\kern-.025em b}\kern-.08em
    T\kern-.1667em\lower.7ex\hbox{E}\kern-.125emX}}

\begin{document}

\title{Decoding the Flow: CauseMotion for Emotional Causality Analysis in Long-form Conversations}

\author{
    \textbf{
        Yuxuan Zhang\textsuperscript{1*}, 
        Yulong Li\textsuperscript{2*\textdagger}, 
        Zichen Yu\textsuperscript{3}, 
        Feilong Tang\textsuperscript{4}, 
        Zhixiang Lu\textsuperscript{5}, 
        Chong Li\textsuperscript{6},
        Kang Dang\textsuperscript{7}
    }\\
    \textbf{Jionglong Su\textsuperscript{8\textdagger}}\\
    \hspace*{-2em} 
    \fontsize{9.5pt}{11.6pt}\selectfont{\textsuperscript{1,2,3,5,6,7,8} School of Artificial Intelligence and Advanced Computing, Xi'an Jiaotong-Liverpool University, Suzhou, 215123, China} \\
    \fontsize{9.5pt}{11.6pt}\selectfont{\textsuperscript{4} Monash University} \\
    \vspace{-1em}
    Yulong.Li19@student.xjtlu.edu.cn, Jionglong.Su@xjtlu.edu.cn
}

\maketitle

\begin{abstract}
Long-sequence causal reasoning seeks to uncover causal relationships within extended time series data but is hindered by complex dependencies and the challenges of validating causal links. To address the limitations of large-scale language models (e.g., GPT-4) in capturing intricate emotional causality within extended dialogues, we propose CauseMotion, a long-sequence emotional causal reasoning framework grounded in Retrieval-Augmented Generation (RAG) and multimodal fusion. Unlike conventional methods relying only on textual information, CauseMotion enriches semantic representations by incorporating audio-derived features—vocal emotion, emotional intensity, and speech rate—into textual modalities. By integrating RAG with a sliding window mechanism, it effectively retrieves and leverages contextually relevant dialogue segments, thus enabling the inference of complex emotional causal chains spanning multiple conversational turns. To evaluate its effectiveness, we constructed the first benchmark dataset dedicated to long-sequence emotional causal reasoning, featuring dialogues with over 70 turns. Experimental results demonstrate that the proposed RAG-based multimodal integrated approach, the efficacy of substantially enhances both the depth of emotional understanding and the causal inference capabilities of large-scale language models. A GLM-4 integrated with CasueMotion achieves an 8.7\% improvement in causal accuracy over the original model and surpasses GPT-4o by 1.2\%. Additionally, on the publicly available DiaASQ dataset, CasueMotion-GLM-4 achieves state-of-the-art results in accuracy, F1 score, and causal reasoning accuracy.
\end{abstract}

\begin{figure}[b]  % [b] 强制放在底部
    \noindent\rule{\linewidth}{0.5pt}  % 添加横线
    \begin{flushleft}
    \footnotesize
    *These authors contributed equally.\\
    \dag Corresponding author.\\
    This is an anonymous repository where we public all the code:
    https://anonymous.4open.science/r/CasueMotion-F463/get\_emo\_no\_rag.py
    \end{flushleft}
\end{figure}

\vspace{-0.5em}
\section{Introduction}
\label{sec:intro}
In natural language processing and affective computing, understanding human emotions and their causal relationships is essential for intelligent human-computer interaction systems \cite{t1,t2,t3,t4}. While Aspect-Based Sentiment Analysis (ABSA) progressed from coarse-grained to fine-grained analysis, with the ability to extract targets, aspects, opinions, and sentiments from text, new challenges have emerged\cite{li-etal-2018-aspect}. The proliferation of social media results in increasingly complex emotional expressions, involving long text sequences and multimodal information such as tone and speech rate. Existing models struggle with extended texts and multimodal data integration, particularly in capturing global context and addressing the forgetting problem in long-sequence emotional causality inference\cite{tian-etal-2020-skep}. These limitations hinder their ability to track emotional causes, development, and impacts effectively\cite{pontiki-etal-2014-semeval}. Thus, there is a growing demand for methods that can process long sequences and multimodal data to better understand complex emotional expressions.

\begin{figure}[t]
    \centering
    \resizebox{0.5\textwidth}{!}{\includegraphics{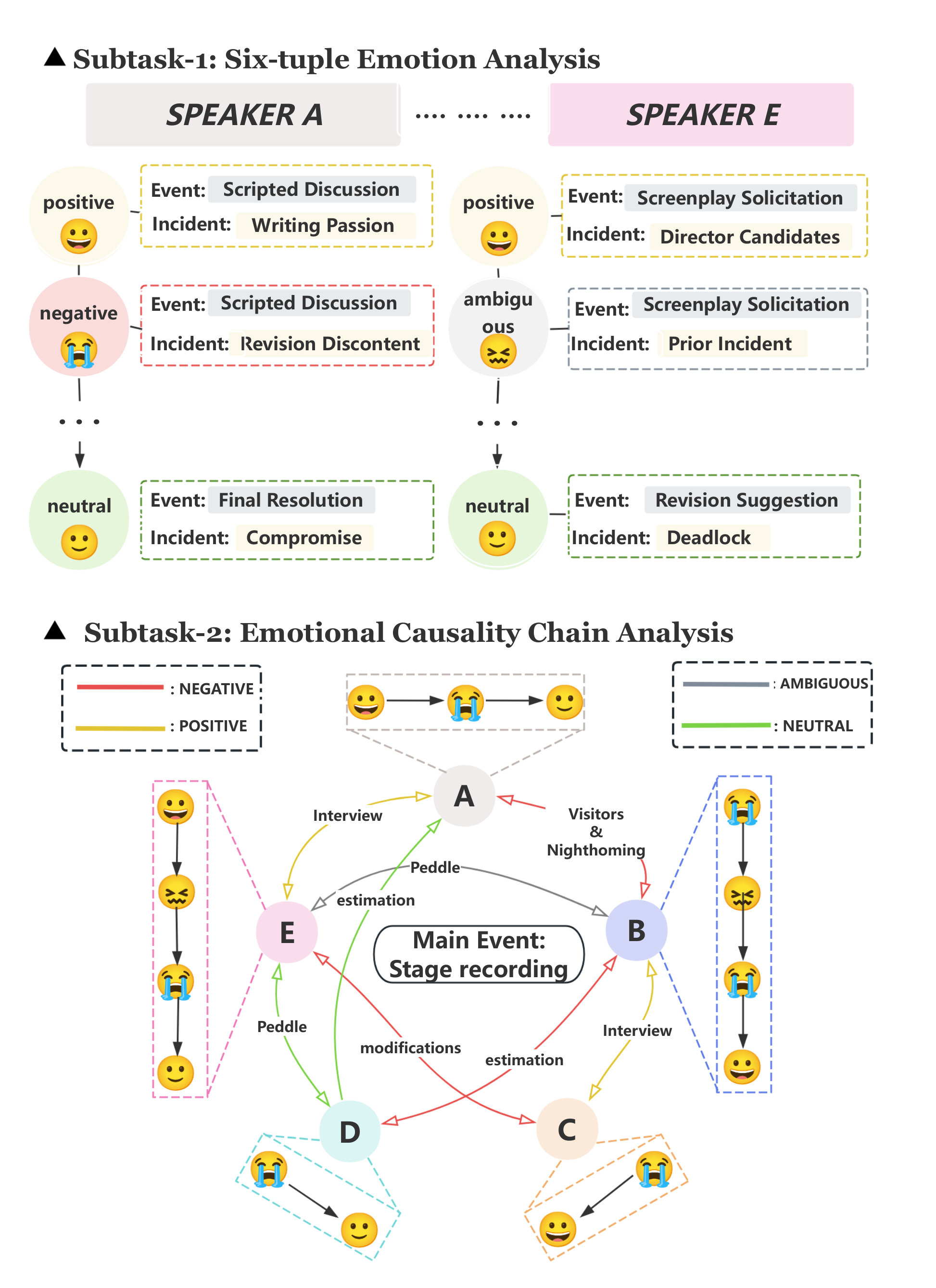}}  % 使用resizebox调整
    \caption{Emotional fluctuation patterns and interpersonal influences during key conversational events among participants. The figure highlights how emotions evolve through various dialogue phases and the effects of interactions such as emotional contagion, support, conflict escalation, and mediation on these emotional dynamics.}
    \label{fig1}
\vspace{-2em}
\end{figure} 
Long-sequence emotional causal reasoning is essential for enabling large-scale language models to conduct complex causal inference in lengthy texts \cite{liao-etal-2021-mmconv}. It seeks to identify the origins, development, and outcomes of emotional states within extended discourse \cite{t9,t10,t11,t12}. However, three primary challenges persist in current research. First, input length constraints necessitate truncating or segmenting text, resulting in the loss of global context and hindering the capture of long-range dependencies across paragraphs or dialogue turns \cite{zhang-etal-2021a-aspect}. Second, modelling long-range dependencies is difficult, impeding the accurate establishment of global causal associations and leading to incomplete or imprecise reasoning \cite{pontiki-etal-2014-semeval}. Third, segment-based processing can disrupt event order and logical relationships, weakening the model’s comprehension of the overall causal chain \cite{li-etal-2023-diasq}. To overcome these challenges, we propose utilizing Retrieval-Augmented Generation (RAG) to build a dialogue knowledge base \cite{lewis-etal-2020-rag}. This enables dynamic retrieval of global context, mitigating input length limitations and enhancing the model’s ability to capture long-range dependencies and complex causal chains.

Multimodal fusion seeks to integrate diverse data types, such as text and speech, to achieve a more comprehensive understanding of affective expressions \cite{ni-etal-2022-recent, li-etal-2022-emocaps}. Audio features—including pitch, speaking rate, and emotional cues—provide essential information for emotion-cause reasoning. However, incorporating audio features into large-scale pre-trained language models presents three major challenges. First, textual and acoustic modalities differ significantly in feature representation and statistical properties; text is inherently discrete, while audio signals are continuous and require complex preprocessing (e.g., Mel spectrograms, MFCC extraction) to align and fuse with textual features \cite{yu2019entity}. Second, the closed-source proprietary nature of large models limits access to their internal structures and parameters, making deep customization or direct integration of audio features difficult \cite{bommasani-etal-2021-opportunities, grattafiori2024llama3herdmodels}. Third, although some large-scale models include speech modules, they often lack the capability for emotion-cause reasoning, hindering effective utilization of audio features in this context \cite{yu2019entity}. To address these challenges, we propose an innovative approach that deeply embeds audio features into the model’s input design and dialogue database, ensuring that acoustic information is continuously integrated throughout the emotion-cause reasoning process.

Current emotional causality inference is limited by existing datasets that impede long-sequence causal reasoning, hindering progress in the field \cite{t5,t6,t7,t8}. Existing datasets face three main challenges. First, these datasets primarily focus on short texts, such as single sentences or brief reviews, lacking the contextual and long-range dependencies essential for long-sequence reasoning \cite{li-etal-2023-diasq}. This restricts models from learning the progression and causal links of emotions in extended texts. Second, although some datasets annotate six key elements, they usually capture only simple, direct causalities and omit complex, multi-level relationships across multiple paragraphs or dialogue turns \cite{luo2024panosent}. This limits the ability to model deep emotional causal chains. Third, many datasets do not support multimodal data \cite{tang-etal-2016-effective}, focusing solely on text and excluding audio features such as pitch, speech rate, or emotional cues. To overcome these limitations, we propose creating a dataset with long sequences, multimodal data, and detailed annotations for complex emotional causal relationships.

We address the challenges of long-sequence emotional causal reasoning by focusing on building comprehensive causal chains to understand complex emotional dynamics. Utilizing a large pre-trained language model, we construct a dataset of 20,000 extended dialogues (70-300 turns) and a validation set of 2,745 authentic dialogues. To handle the limitations of long texts, we integrate RAG technology with a dialogue knowledge base, enabling dynamic retrieval of global context and precise modelling of causal links across extensive dialogues. Additionally, audio features—pitch, speech rate, and emotional state—are embedded into the input design and database, enriching the causal reasoning process.

Our main contributions are as follows:
\begin{itemize}
\item We propose a method for integrating audio features into both the model’s input and the dialogue knowledge base. By embedding audio information within the knowledge base, the model simultaneously leverages textual and auditory data for emotion-cause reasoning, enabling effective multimodal fusion.

\item We provide the largest dataset for long-sequence emotional causality in dialogues. This dataset addresses key gaps in emotion cause-effect inference by providing hierarchical multi-turn conversations with fine-grained emotion cause-effect pairs and substantial dialogue depth.

\item We propose CauseMotion, a novel causal reasoning framework that integrates RAG into the inference process, effectively capturing long-range dependencies and complex causal chains. CauseMotion achieves state-of-the-art performance on the DiaASQ dataset, and the CauseMotion-GLM-4 variant outperforms GPT-4o across all metrics on the ATLAS dataset.
\end{itemize}
\vspace{-0.5em}
\section{DataSet}
We construct and utilize ATLAS-6, a dataset for long-sequence emotional causal reasoning that consists of two parts: an auxiliary synthetic dataset and a real-world validation dataset. The auxiliary synthetic dataset consists of 20,000 extended dialogue texts (ranging from 70 to 300 turns), covering eight different scenarios including customer service, social media interactions, medical consultations, educational dialogues, technical support, emotional support, market research, and multi-party discussions. These dialogues, assisted by large language model (LLMs), simulate complex emotional causal relationships and long-range dependencies typically found in real-world contexts. After data generation, each utterance is rigorously annotated with six key elements—Holder, Target, Aspect, Opinion, Sentiment, and Rationale—ensuring comprehensive and fine-grained labeling.

 \begin{figure}[t]
    \centering
    \includegraphics[trim=0 0 0 50, clip, width=0.5\textwidth]{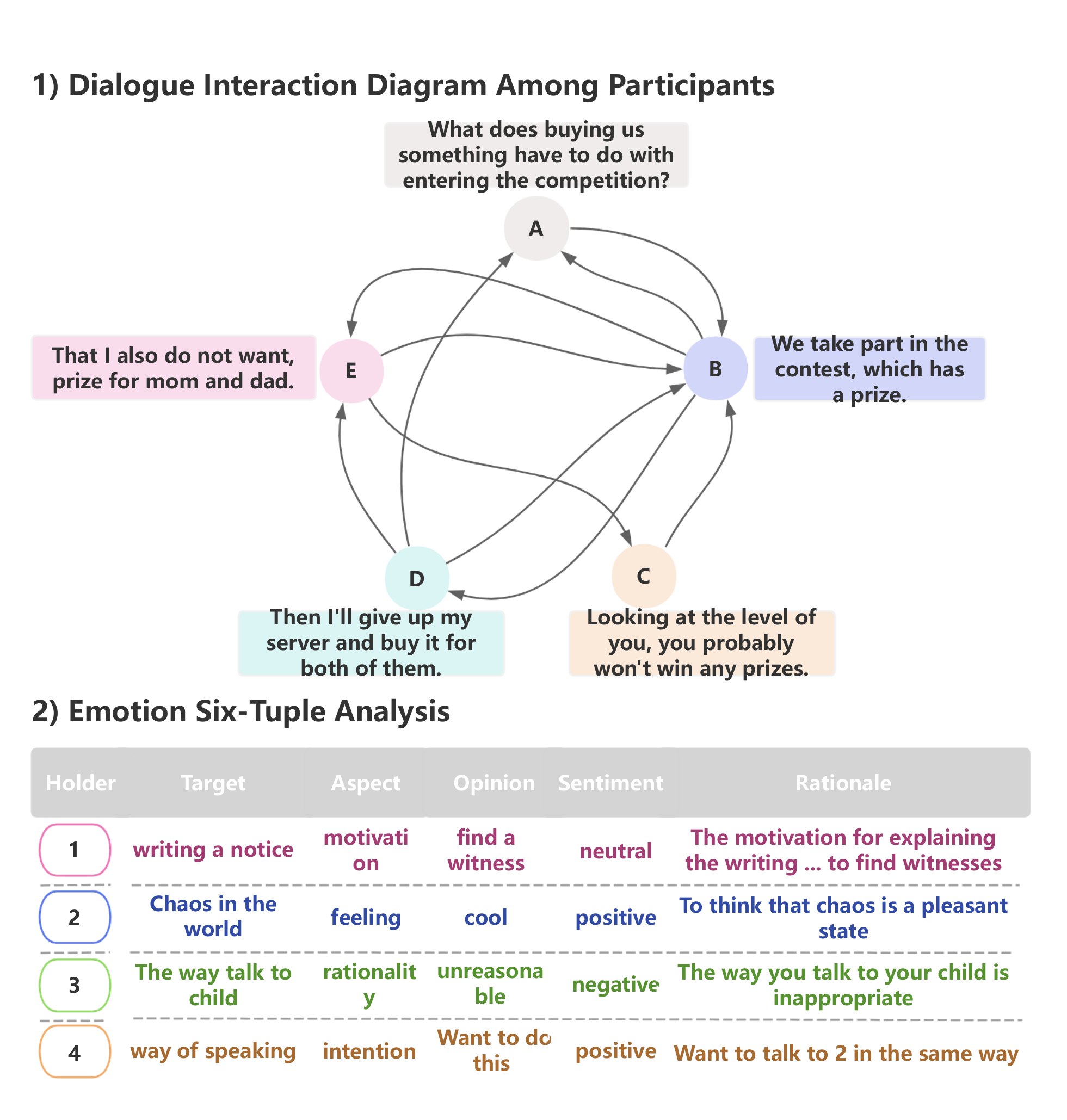}  % 使用resizebox调整
    \vspace{-2.5em}
    \caption{Interactional patterns among interlocutors and the corresponding emotional dynamics of a specific participant within the dialogue. The figure maps various interaction types—including supportive exchanges, confrontational dialogues, and neutral statements—to the participant's emotional states such as happiness, frustration, anger, and calmness.}
    \setlength{\textfloatsep}{30pt}
    \label{fig2}
\vspace{-1em}
\end{figure}

The real-world validation dataset contains 2,745 long-sequence dialogues derived from media such as film and social networks, also spanning 70 to 300 turns. As shown in Fig.\ref{fig2}, these dialogues authentically reflect emotional expressions and causal relationships across multiple interaction rounds. The real-world dialogues undergo stringent manual annotation and cross-checking procedures.

Taken together, these two datasets provide a foundational resource for training and evaluating models in diverse and complex long-sequence scenarios. This approach not only addresses the gaps in existing datasets, which often lack extensive long-range dependency and detailed causal relationship annotations but also enables models to exhibit superior generalization capabilities in real-world contexts\cite{pontiki-etal-2014-semeval}. By offering richly annotated long-sequence dialogues, the study lays a solid groundwork for developing models capable of deeper emotional understanding and causal inference, thereby advancing research in multimodal emotional analysis and intelligent dialogue systems toward more human-level emotional intelligence.

\begin{figure*}[ht]
	\centering
	\includegraphics[trim=0 30 0 30, clip, width=2\columnwidth]{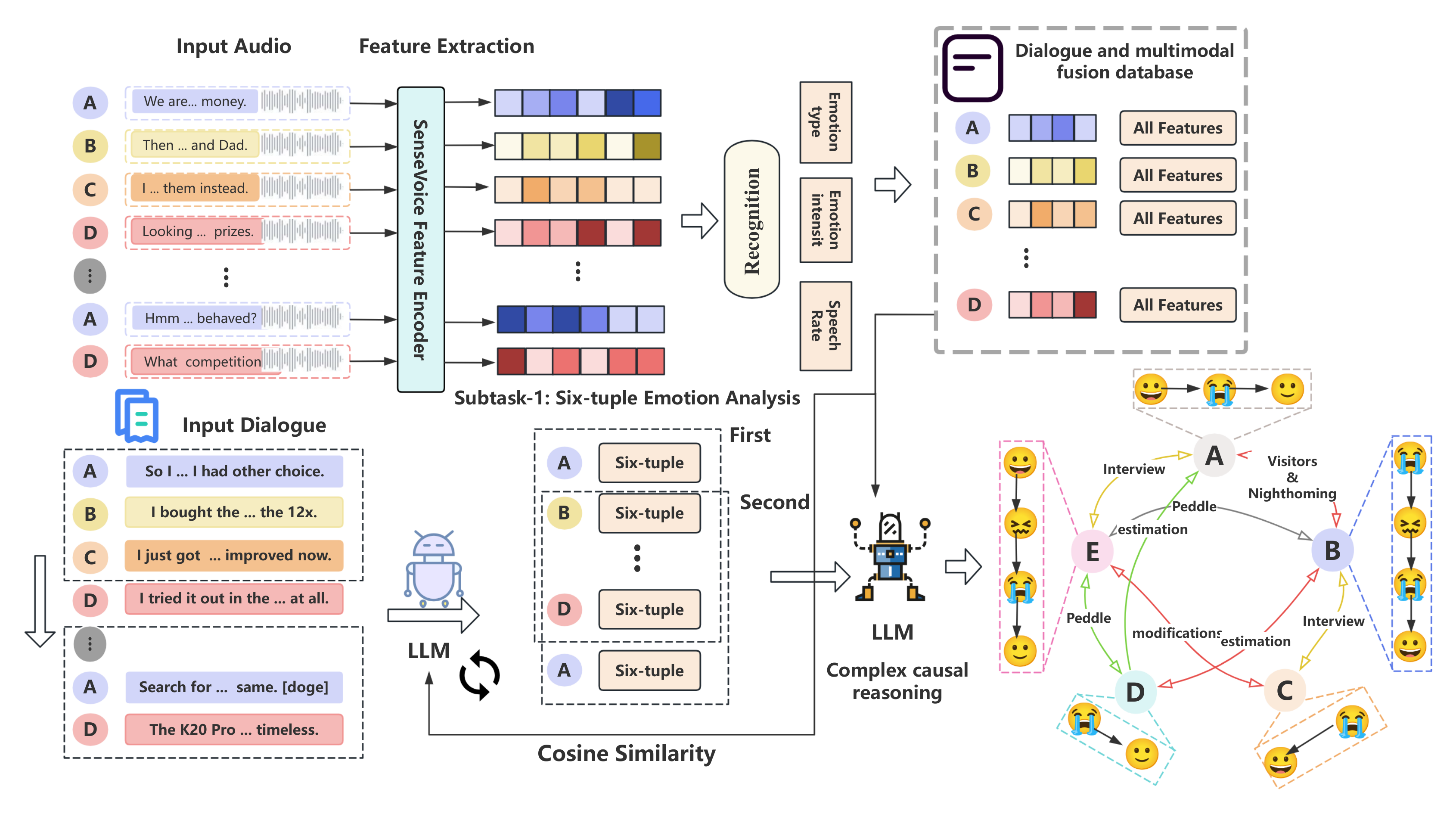} 
	\caption{An overview of our multimodal emotion analysis framework. The framework first extracts audio and dialogue features using the SoundVoice Feature Encoder for six-tuple emotion analysis, encompassing Holder, Target, Aspect, Opinion, Sentiment, and Rationale. The Recognition Module integrates these features, followed by an LLM-based causal reasoning component to analyze emotional relationships. The output is visualized as a social interaction graph, illustrating the emotional dynamics and interactions between participants.}
	\label{fig3}
\vspace{-1em}
\end{figure*}

The auxiliary synthetic dataset comprises 20,000 long-sequence dialogue texts generated with assistance from LLMs. Each dialogue contains 70 to 300 turns and covers eight distinct application scenarios: customer service, social media interactions, medical consultations, educational dialogues, technical support, emotional support, market research, and multi-party discussions. As shown in Fig.\ref{fig1}, these synthetic dialogues simulate complex emotional causal relationships and long-range dependencies found in real conversations, ensuring data diversity and complexity. We implement a rigorous annotation verification process, annotating each dialogue for emotional causality with six key elements: Holder, Target, Aspect, Opinion, Sentiment, and Rationale. The data is annotated by multiple professionally trained professional researchers, and the Sentiment scores are averaged after being independently rated by the experts to ensure accuracy and consistency in the annotations.

\section{Methodology}
To address the technical challenges in long-sequence emotional causality inference, we propose a comprehensive solution. We define the input data as a dialogue consists \textit{n} sets \( D = \{u_1, u_2, \ldots, u_n\} \), where each utterance \( u_i = \{w_{i1}, w_{i2}, \ldots, w_{im}\} \) consists of \( m \) words. To capture long-range dependencies and complex causal chains, we employ a sliding time window approach, creating local dialogue subsets \( D_t = \{u_t, u_{t+1}, \ldots, u_{t+k}\} \), where \( t \) is the current time step and \( k \) is the window size. By continuously sliding the window, we construct a dialogue knowledge base \( K = \{D_{t1}, D_{t2}, \ldots, D_{tm}\} \) that includes multiple local dialogue subsequences, each reflecting contextual information and causal relationship patterns at different time steps.

In the causal inference process, we introduce the Retrieval-Augmented Generation (RAG) technology. Specifically, when the model receives the current input dialogue \( D_t \), the RAG module retrieves the most relevant dialogue subsequences \( C_t \) from the dialogue knowledge base \( K \). These retrieved contextual information \( C_t \) are input into the generation model together with the current input \( D_t \) to assist in extraction. The objective of the model is to extract all possible emotional causality sextuplets \( Q = \{{(h_j, t_j, a_j, o_j, p_j, r_j)}\}_{j=1}^K \) from the input time window \( W \),
where \( h_k \) means the Holder, \( t_k \) means the Target, \( a_k \) means the Aspect, \( o_k \) means the Opinion, \( p_k \) means the Sentiment, and \( r_k \) means the rationale.
\( p_k \) is the label of the sentiment category, and \( r_k \) is the basis for the inference of emotional causality.

\subsection{Multimodal Fusion Mechanism}
To further enhance model performance in long-sequence emotional causality inference tasks, we design a multimodal fusion mechanism that deeply integrates audio features into the dialogue knowledge base and prompts. Specifically, we use SenseVoice\cite{sense} to extract emotional features from the audio, including emotion in the voice and emotional intensity, and mathematically calculate the speech rate. As shown in Fig.\ref{fig3}, these audio features are then converted into textual descriptions and combined with dialogue text information, achieving effective multimodal data fusion and significantly improving the accuracy and depth of emotional causality inference.

Key audio-emotional features are extracted from speech data in dialogues. Using SenseVoice to process audio signals, we extract a set of audio features including voice emotion $ e_i \in \mathbb{R}^d $ and emotional intensity $ \theta_i \in \mathbb{R} $, where each audio feature vector $ a_i $ is defined as:
\vspace{-0.5em}
\begin{equation}
    a_i = \{e_i, \theta_i\},
\end{equation}
where $ e_i $ denotes the voice emotion feature vector of the $ i $th audio segment with dimension $ d $, and $ \theta_i $ represents the emotional intensity value of $a_i$.

Additionally, speech rate is defined as the number of words per unit of time and calculated as:
\vspace{-0.5em}
\begin{equation}
    r_i = \frac{m}{t_i^{\text{end}} - t_i^{\text{start}}},
\end{equation}
where $ m $ is the number of words in utterance $ u_i $, $ t_i^{\text{end}} $ and $ t_i^{\text{start}} $ describe the end and start times of that utterance, respectively. The resulting speech rate feature set is $ R = \{r_1, r_2, \ldots, r_n\} $, where each $ r_i \in \mathbb{R} $.

 The text embedding matrix be $ E_t \in \mathbb{R}^{d_t} $, the audio emotion embedding matrix be $ E_e \in \mathbb{R}^{d_e} $, and the speech rate embedding matrix be $ E_r \in \mathbb{R} , \mathbb{R}$ 
 means real number, where \( d_t \) denotes the dimensionality of the text embeddings and \( d_e \) means the dimensionality of the audio emotion embeddings. The multimodal embedding $ E_m $ can then be represented as:
\vspace{-0.5em}
\begin{equation}
    E_m = \text{Concat}(E_t, E_e, E_r),
\end{equation}
where $\text{Concat}$ defines the concatenation operation of text embedding $ E_t $, audio emotion embedding $ E_e $, and speech rate embedding $ E_r $, forming a fused feature vector $ e_i \in \mathbb{R}^{d_m}, d_m=d_t+d_e+1 $.

The fused multimodal embedding \( E_m \) is integrated into the dialogue knowledge base \( K_d \) and the generative model’s prompts. It enhances input prompts with auxiliary information to improve emotional context understanding during sextuplet generation. Each time window \( W_j \) in \( K_d \) includes text dialogues and multimodal features \( E_{mj} \), enabling context retrieval enriched with emotional cues:
\vspace{-0.5em}
\begin{equation}
K_d = \{(W_1, E_{m1}), (W_2, E_{m2}), \ldots, (W_j, E_{mj})\},
\end{equation}
where $ W_j $ represents the dialogue segment of the $ j $th time window and $ E_{mj} $ means the corresponding multimodal feature vector.

\begin{table*}[t!]
    \centering
    \caption{Comparison with the SOTA CauseMotion models and other causal relation extraction approaches on the DiaASQ dataset. Best results are in bold. }
    \vspace{-0.5em}
    \begin{tabular*}{1\textwidth}{@{\extracolsep{\fill}} l l ccc ccc }
        \hline
        & \multicolumn{3}{c}{Span Match (F1)} & \multicolumn{3}{c}{Pair Extraction (F1)}  \\
        \cline{2-4}
        \cline{5-7}
        & T & A & O & T-A & T-O & A-O  \\ 
        \hline
        \multirow{6}{*} 
        \text{CRF-Extract-Classify}\cite{cai-etal-2021-aspect}     & 91.11 & 75.24 & 50.06 & 32.47 & 26.78 & 18.90  \\
        SpERT\cite{eberts-ulges-2020-span}                           & 90.69 & 76.81 & 54.06 & 38.05 & 31.28 & 21.89  \\
        DiaASQ\cite{li-etal-2023-diasq}                          & 90.23 & 76.94 & 59.35 & 48.61 & 43.31 & 45.44  \\
        ParaPhrase\cite{zhang-etal-2021a-aspect}                     & / & / & / & 37.81 & 34.32 & 27.76  \\
        Span-ASTE\cite{xu-etal-2021-learning}                       & / & / & / & 44.13 & 34.46 & 32.21  \\
        \hline
        CauseMotion-LLama-3.3-70B        & 91.21 & \textbf{77.66} & 60.34 & 63.98 & 50.09 & 58.63 \\
        CauseMotion-internLM2\_5-20b    & 90.18 & 75.33 & 58.12 & 59.21 & 45.66 & 55.34  \\
        CauseMotion-Qwen2.5-72B        & 91.22 & 76.93 & 59.89 & 63.02 & 49.78 & 57.12 \\
        \textbf{CauseMotion-GLM-4}               & \textbf{91.43} & 77.63 & \textbf{61.35} & \textbf{64.15} & \textbf{50.22} & \textbf{59.16} \\
        \hline
    \end{tabular*}
    \vspace{-1em}
    \label{table_1}
\end{table*}

\subsection{Construction of Dialogue Knowledge Base and Long-sequence RAG}

We apply RAG technology to retrieve the most relevant dialogue segments $ C_j $ from the dialogue knowledge base $ K_d $. This retrieval process can be formally defined as:
\vspace{-0.5em}
\begin{equation}
C_j = \text{RAG}(W_j, K_d)    ,
\end{equation}
where $ C_j $ denotes the set of dialogue segments semantically related to the current time window $ W_j $. 

The RAG module calculates the similarity between the current time window $ W_j $ and each time window $ W_i $ in the knowledge base, selecting several time windows with the highest similarity as retrieval results. The similarity is calculated using the cosine similarity formula:
\vspace{-0.5em}
\begin{equation}
\text{Similarity}(W_j, W_i) = \frac{W_j \cdot W_i}{\|W_j\| \|W_i\|},
\end{equation}
where $ W_j $ and $ W_i $ describe the vector representations of the current time window $W_j$ and the $ i $th time window in the knowledge base, respectively. The dot product \( {W_j} \cdot {W_i} \) means the inner product of vectors \( {W_j} \) and \( W_i \).

 The retrieved contextual information $ C_j $ is input into the generative model along with the current input $ W_j $ to assist in extracting sextuplets $ Q_j = \{(h_k, t_k, a_k, o_k, p_k, r_k)\} $ and causal reasoning. 
\subsection{Complex Causal Chain Reasoning}
\noindent A causal connection is established based on three metrics: Senmantic\_Score, Temporal\_Score and Rationale\_Score.\\

1) Semantic consistency measures the alignment between the opinion \( o_{j_k} \) in \( Q_j \) and the sentiment \( p_{i_k} \) in \( Q_i \) using:
 \vspace{-0.5em}
\begin{equation}
\text{Semantic\_Score}(o_{j_k}, p_{i_k}) = \frac{o_{j_k} \cdot p_{i_k}}{\|o_{j_k}\| \|p_{i_k}\|},
\end{equation}

2) Temporal constraints verify the causality based on the time gap \( \Delta t_{ij} \) as follows:
\vspace{-0.5em}
\begin{equation}
\Delta t_{ij} = t_{i_k}^{\text{start}} - t_{j_k}^{\text{end}},
\end{equation}
\vspace{-2em}
\begin{equation}
     \quad \text{Temporal\_Score}(\Delta t_{ij}) = \exp\left(-\frac{\Delta t_{ij}}{\tau}\right),
\end{equation}
where \( \Delta t_{ij} \) quantifies the temporal gap between the two events, \( t_{j_k}^{\text{end}} \) is the end time of the event in \( Q_j \), and \( t_{i_k}^{\text{start}} \) is the start time of the event in \( Q_i \). \( \tau \) is the maximum allowable time gap. 

3) Rationale alignment evaluates the logical support provided by the rationale \( r_{j_k} \) in \( Q_j \) for \( Q_i \), assessed via a probability score \( P_{\text{NLI}} \) derived from natural language inference (NLI) models:
\vspace{-0.5em}
\begin{equation}
\text{Rationale\_Score}(r_{j_k}, Q_i) = \log\left(1 + P_{\text{NLI}}(r_{j_k} \rightarrow Q_i)\right)
\end{equation}
where \( P_{\text{NLI}}(r_{j_k} \rightarrow Q_i) \) represents the probability that the rationale \( r_{j_k} \) logically supports \( Q_i \). 

Based on the three metrics, the causal relationships are represented as a directed graph \( G = (V, E) \), where edges \( e_{ij} \) are weighted as:
\vspace{-0.6em}
\begin{equation}
\begin{split}
    \text{Weight}(e_{ij}) = & \alpha \cdot \text{Semantic\_Score} +\beta \cdot \text{Temporal\_Score}\\
                            & + \gamma \cdot \text{Rationale\_Score}
\end{split}
\end{equation}
\vspace{-0.5em}

\noindent\text{where $\alpha$, $\beta$ and $\gamma$are weight parameters and $\alpha + \beta + \gamma =1$}
$(0< \alpha, \beta,\gamma< 1) $ .

\begin{table*}[t]
\vspace{-0.5em}
\renewcommand{\arraystretch}{}
\centering
\caption{Ablation results (Micro F1). Comparison of model performance with and without CauseMotion implementation across different evaluation metrics. The best results are in bold.}
\vspace{-0.5em}
\begin{tabular}{l l ccccc c cc}
\hline
\multirow{10}{*}{} & & Target & Aspect & Opinion & Rationale & Sentiment & & Causality Chain Accuracy \\
\cline{2-7} \cline{9-9}
 & GPT-4o  & 0.450 & 0.342 & 0.541 & 0.723 & 0.674 & & 0.562 \\
\hline
 & LLama-3.3-70B & 0.394\scriptsize{$_{\downarrow0.056}$} & 0.305\scriptsize{$_{\downarrow0.037}$} & 0.454\scriptsize{$_{\downarrow0.087}$} & 0.578\scriptsize{$_{\downarrow0.145}$} & 0.585\scriptsize{$_{\downarrow0.089}$} & & 0.444\scriptsize{$_{\downarrow0.118}$} \\
 & internLM2\_5-20b & 0.374\scriptsize{$_{\downarrow0.076}$} & 0.252\scriptsize{$_{\downarrow0.090}$} & 0.383\scriptsize{$_{\downarrow0.158}$} & 0.460\scriptsize{$_{\downarrow0.263}$} & 0.518\scriptsize{$_{\downarrow0.156}$} & & 0.409\scriptsize{$_{\downarrow0.153}$} \\
 & Qwen2.5-72B & 0.355\scriptsize{$_{\downarrow0.095}$} & 0.266\scriptsize{$_{\downarrow0.076}$} & 0.452\scriptsize{$_{\downarrow0.089}$} & 0.584\scriptsize{$_{\downarrow0.139}$} & 0.560\scriptsize{$_{\downarrow0.114}$} & & 0.421\scriptsize{$_{\downarrow0.141}$} \\
 & GLM-4 & 0.418\scriptsize{$_{\downarrow0.032}$} & 0.332\scriptsize{$_{\downarrow0.010}$} & 0.474\scriptsize{$_{\downarrow0.067}$} & 0.585\scriptsize{$_{\downarrow0.138}$} & 0.579\scriptsize{$_{\downarrow0.095}$} & & 0.487\scriptsize{$_{\downarrow0.075}$} \\
\cdashline{1-9}[1pt/2pt]
\multirow{4}{*}{\textbf{CauseMotion}} & LLama-3.3-70B & 0.406\scriptsize{$_{\downarrow0.044}$} & 0.304\scriptsize{$_{\downarrow0.038}$} & 0.424\scriptsize{$_{\downarrow0.117}$} & 0.656\scriptsize{$_{\downarrow0.067}$} & 0.589\scriptsize{$_{\downarrow0.085}$} & & 0.461\scriptsize{$_{\downarrow0.101}$} \\
 & InternLM2\_5-20b & 0.422\scriptsize{$_{\downarrow0.028}$} & 0.307\scriptsize{$_{\downarrow0.035}$} & 0.384\scriptsize{$_{\downarrow0.157}$} & 0.678\scriptsize{$_{\downarrow0.045}$} & 0.640\scriptsize{$_{\downarrow0.034}$} & & 0.505\scriptsize{$_{\downarrow0.057}$} \\
 & Qwen2.5-72B & 0.275\scriptsize{$_{\downarrow0.175}$} & 0.282\scriptsize{$_{\downarrow0.060}$} & 0.477\scriptsize{$_{\downarrow0.064}$} & 0.681\scriptsize{$_{\downarrow0.042}$} & 0.611\scriptsize{$_{\downarrow0.063}$} & & 0.474\scriptsize{$_{\downarrow0.088}$} \\
 & \textbf{GLM-4} & \textbf{0.523\scriptsize{$_{+0.073}$}} & \textbf{0.453\scriptsize{$_{+0.111}$}} & \textbf{0.592\scriptsize{$_{+0.051}$}} & \textbf{0.731\scriptsize{$_{+0.008}$}} & \textbf{0.724\scriptsize{$_{+0.050}$}} & & \textbf{0.574\scriptsize{$_{+0.012}$}} \\
\hline
\vspace{-1em}
\end{tabular}
\label{tab:2}
\end{table*}

\vspace{-0.5em}
\section{Experiment}
\vspace{-0.5em}
We experiment with five large models, including open-source models LLama-3.3\cite{grattafiori2024llama3herdmodels}, Qwen2.5-72B\cite{bai2023qwentechnicalreport} , InternLM2\_5-20b\cite{cai2024internlm2technicalreport}, as well as proprietary models GLM-4\cite{glm2024chatglmfamilylargelanguage} and GPT-4o\cite{openai2024gpt4technicalreport}. For open-source models, we employ distributed training across 64 A800 GPUs, while proprietary models were accessed through their official APIs.

\subsection{Evaluation Metrics}
We employ three metrics to evaluate emotion causality reasoning: Causal Correctness, Causal Consistency, and the combined Causal Chain Score.

\textit{Correct Causal Links} is the number of predicted causal relationships that align with the annotated ground-truth causal connections, and \textit{Total Predicted Causal Links} is the total number of causal relationships identified by the model.
\begin{equation}
\text{Causal\_Correctness} = \frac{\text{Correct Causal Links}}{\text{Total Predicted Causal Links}}
\end{equation}
% where \textit{Correct Causal Links} is the number of predicted causal relationships that align with the annotated ground-truth causal connections, and \textit{Total Predicted Causal Links} is the total number of causal relationships identified by the model.

\textit{Consistent Causal Links} denotes the number of logically coherent causal relationships in the prediction, and \textit{Total Causal Links} represents the entire set of inferred causal connections.
\begin{equation}
\text{Causal\_Consistency} = \frac{\text{Consistent Causal Links}}{\text{Total Causal Links}}
\end{equation}

\textit{Causal Correctness} and \textit{Causal Consistency} are weighted equally (0.5 each), ensuring that both the accuracy of the predicted links and their logical coherence contribute equally to the final score.
\begin{equation}
\begin{split}
\text{Causal\_Chain\_Score} = & 0.5 \cdot \text{Causal\_Correctness} \\
                              & + 0.5 \cdot \text{Causal\_Consistency}
\end{split}
\end{equation}

\begin{figure}[h]
    \centering
    \includegraphics[width=1\linewidth]{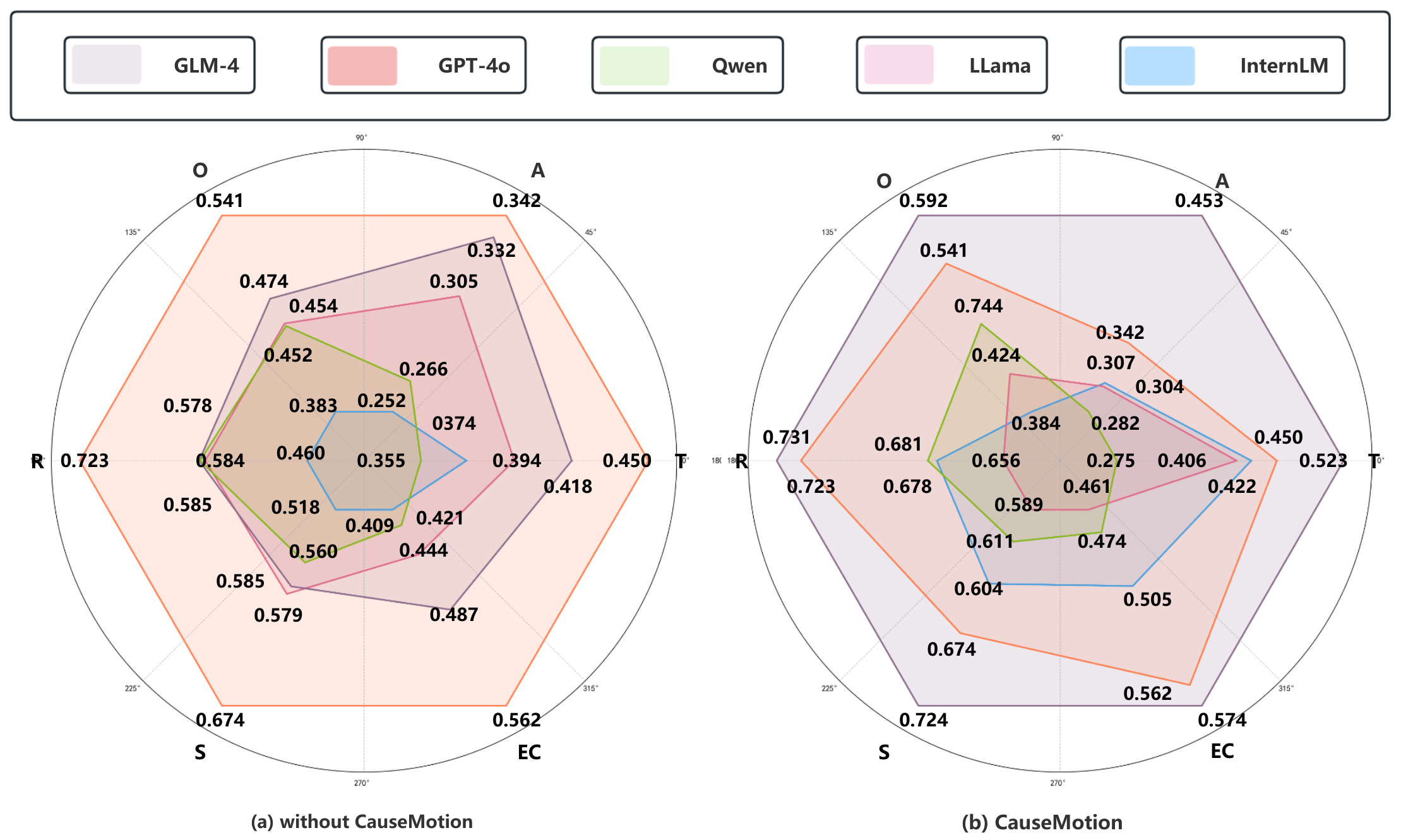}
    \caption{The average performance is evaluated on 2,745 real-world samples from the ATLAS-6 dataset.}
    \label{fig:RAG}
\vspace{-1.5em}
\end{figure}

\subsection{Result}
% We evaluated various causal relation extraction models on the \textbf{DiaASQ} dataset TABLE \ref{table_1}. The \textbf{CauseMotion-GLM-4} model outperforms all other models in both \textit{Span Match (F1)} and \textit{Pair Extraction (F1)}. Specifically, it achieves \textbf{91.43} for Target span matching, \textbf{77.63} for Aspect span matching, and \textbf{61.35} for opinion.

% On the ATLAS dataset, as shown in TABLE \ref{tab:2} and Fig. \ref{fig:RAG}, the CauseMotion framework significantly enhances Emotional Causality Reasoning Chain Accuracy, demonstrating its superiority in capturing long-range dependencies and complex causal relationships. The CauseMotion-GLM4 model achieves the best performance with Emotional Causality Reasoning Chain Accuracy of 0.574 outperforming GPT-4o and other models. Without CauseMotion, the performance drops substantially, highlighting the critical role of our framework in improving emotional causality reasoning. 

% 1\\
% 2\\
% 3\\

We evaluated various causal relation extraction models on the \textbf{DiaASQ} dataset, as shown in Table \ref{table_1}. The \textbf{CauseMotion-GLM-4} model significantly outperforms all other models in two key evaluation metrics: \textit{Span Match (F1)} and \textit{Pair Extraction (F1)}. Specifically, it achieves \textbf{91.43} in Target span matching F1 score, \textbf{77.63} in Aspect span matching F1 score, and \textbf{61.35} in Opinion extraction F1 score. These results highlight the superior accuracy of CauseMotion-GLM-4 in identifying and extracting causal relationships across different aspects of the data.

On the ATLAS dataset, as shown in Table \ref{tab:2} and Fig.\ref{fig:RAG}, the CauseMotion framework demonstrates a significant improvement in Emotional Causality Reasoning Chain Accuracy. The CauseMotion-GLM-4 model achieves the highest performance with an Emotional Causality Reasoning Chain Accuracy of 0.574, outperforming GPT-4o and other state-of-the-art models. Specifically, GPT-4o achieves an accuracy of 0.528, highlighting a clear 8.7\% improvement in accuracy by incorporating the CauseMotion framework.

Without the CauseMotion framework, the performance drops substantially, underscoring the critical role of our approach in improving emotional causality reasoning. This experimental result demonstrates the effectiveness of CauseMotion in processing complex emotional relationships in long dialogues and conducting emotional reasoning analysis
\vspace{-0.5em}
\section{Conclusion}
\vspace{-0.7em}
In this work, we propose CauseMotion, an innovative framework for emotional causality reasoning that integrates RAG technology to enhance long-sequence inference. On the ATLAS dataset, CauseMotion demonstrates significant improvements in capturing long-range dependencies and complex causal chains. The CauseMotion-GLM-4 model achieves state-of-the-art performance, consistently outperforming GPT-4o across all evaluation metrics. By aligning predictions with ground truth annotations while maintaining internal coherence, CauseMotion establishes a new benchmark for emotional causality reasoning. Future research will focus on extending this framework to additional domains and incorporating multimodal data to further enhance its applicability and robustness.

\bibliographystyle{IEEEbib}
\vspace{-0.5em}
\bibliography{references}
\end{document}